# Feature Embedding Clustering using POCS-based Clustering Algorithm

Le-Anh Tran, Dong-Chul Park
Dept. of Electronics Engineering, Myongji University, Yongin, South Korea.

**Abstract:** *An application of the POCS-based clustering algorithm (POCS stands for Projection Onto Convex Set), a novel clustering technique, for feature embedding clustering problems is proposed in this paper. The POCS-based clustering algorithm applies the POCS's convergence property to clustering problems and has shown competitive performance when compared with that of other classical clustering schemes in terms of clustering error and execution speed. Specifically, the POCS-based clustering algorithm treats each data point as a convex set and applies a parallel projection operation from every cluster prototype to corresponding data members in order to minimize the objective function and update the prototypes. The experimental results on the synthetic embedding datasets extracted from the 5 Celebrity Faces and MNIST datasets show that the POCS-based clustering algorithm can perform with favorable results when compared with those of other classical clustering schemes such as the K-Means and Fuzzy C-Means algorithms in feature embedding clustering problems.*

## 1. INTRODUCTION

Along with classification, clustering is one of the most fundamental tasks in automation systems deployed using machine learning [1][2][3][4]. Even though classification and clustering have certain similarities, the difference is that classification is adopted with predefined classes or labeled data and is geared towards supervised learning, while clustering is an unsupervised data analysis technique that aims to analyze data points in order to segregate groups with similar traits and assign them into clusters [5][6]. Popular clustering schemes try to find homogeneous subgroups that have similar characteristics based on the employed objective measure.

One of the most popular and classical methods for general clustering tasks is the K-Means clustering algorithm, which applies the Euclidean distance to measure the similarities among data points [5]. The K-Means clustering algorithm alternates between assigning cluster membership for each data point to the nearest cluster center and computing the center of each cluster as the prototype of its member data points [5]. The purpose of the K-Means clustering algorithm is to obtain a set of cluster prototypes that can minimize the pre-defined objective function. The training procedure of the K-Means clustering algorithm is terminated when there is no further update in the assignment of instances to clusters [5]. The convergence of the K-Means clustering algorithm considerably depends on the initial prototypes and data presentation sequence. However, there exists no efficient and universal method for identifying the initial partitions [6]. In addition, the K-Means algorithm is known to be sensitive to noise and outliers [5].

Another classical clustering algorithm that has also been applied to various clustering tasks is the Fuzzy C-Means (FCM) clustering algorithm [7]. Unlike the K-Means algorithm, in the FCM algorithm, a data point can belong to multiple subgroups concurrently. A membership function is used to represent the degree of certainty for a data point belonging to a certain cluster. The performance of the FCM algorithm is also highly dependent on the selection of the initial prototypes and the initial membership value [7]. Moreover, the disadvantages of the FCM clustering algorithm include extended computational time and incapability in handling noisy data and outliers [7]. In order to handle the computational complexity and upgrade the convergence speed of the FCM algorithm, Park and Dagher introduced the Gradient-Based Fuzzy C-Means (GBFCM) algorithm [8] which combines the FCM algorithm and the characteristics of Kohonen's Self Organizing Map [9] to improve performance.

On the other hand, the projection onto convex set (POCS) method is a robust tool for various purposes such as signal synthesis and image restoration which was introduced by Bregman in the mid-1960s [10]. The POCS method has been widely applied to find a common point of convex sets in several signal processing problems. The aim of the POCS method is to find a vector that resides in the intersection of convex sets. Bregman has shown that successive projections between two or more convex sets with non-empty intersections converge to a point that exists in the intersection of the convex sets [10]. When the convex sets are disjoint, the sequential projection converges to greedy limit cycles which are dependent on the order of the projections instead of converging to a single point [10]. This convergence property of the POCS method has been applied to clustering problems and the POCS-based clustering algorithm [11] has been proposed in our preliminary work [11] which has been proved to be able to perform competitively when compared with other classical clustering schemes such as the K-Means and FCM algorithms. The POCS-based clustering algorithm treats each data point as a convex set and projects the prototypes of the clusters to each of its constituent instances to minimize the objective function and compute a new set of center points. In this paper, we further examine the applicability of the POCS-based clustering

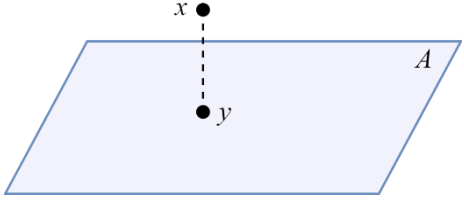

Fig. 1. The projection of x onto convex set A is the unique point in A which is closest to x and is denoted as y.

algorithm to complex clustering tasks such as feature embedding clustering.

The rest of this paper is structured as follows. Section 2 briefly reviews the POCS method and the POCS-based clustering algorithm. The procedure of preparing feature embedding datasets is presented in Section 3. In Section 4, the performance of the POCS-based clustering algorithm on the prepared feature embedding datasets is examined and compared with that of other classical clustering approaches including the K-Means and FCM algorithms. Section 5 concludes the paper.

## 2. POCS-BASED CLUSTERING ALGORITHM

### 2.1 Convex Set

Convex set has been one of the most classical and powerful tools in the optimization theory [10]. A set of data points is called a convex set if it has the following property: given a non-empty set $A$ which is the subset of a Hilbert space $H$, $A \subseteq H$ is called convex, for $\forall x_1, x_2 \in A$ and $\forall \lambda \in [0, 1]$, if the following holds true:

$$x := \lambda x_1 + (1 - \lambda)x_2 \in A \qquad (1)$$

Note that if $\lambda = 1$, $x = x_1$, and if $\lambda = 0$, $x = x_2$. For any value of $0 \leq \lambda \leq 1$ and $x \in A$, x lies on the line segment joining $x_1$ and $x_2$ when the set is convex.

### 2.2 Projection onto Convex Set

The concept of the projection of a point to a plane is utilized to solve many optimization problems such as finding a point on the plane that has the minimum distance from the center of projection. For a given point $x \notin A$, the projection of x onto $A$ is the unique point $y \in A$ such that the distance between x and y is minimum. If $x \in A$, the projection of x onto $A$ is x. The optimization task can be expressed as:

$$y = argmin\|x - y^*\|^2 \qquad (2)$$

where $y^*$ is all the points on the convex set $A$. A graphical illustration of the projection of a point onto a convex set is shown in Fig. 1.

### 2.3 Parallel Projection onto Convex Sets

In the parallel mode of the POCS method, a point is projected to all convex sets concurrently. All the projections are combined convexly with corresponding weight values to solve the minimization problem. For a set of $n$ convex sets $C = \{c_i | 1 \leq i \leq n\}$, the weighted simultaneous projections can be computed as follows:

$$x_{k+1} = x_k + \sum_{i=1}^{n} w_i(P_{c_i} - x_k), \qquad k = 0,1,2,\dots \quad (3)$$

$$\sum_{i=1}^{n} w_i = 1 \qquad (4)$$

where $P_{c_i}$ is the projection of $x_k$ onto convex set $c_i$ and $w_i$ is the weight of importance of the projection. Note that $x_k$ represents the $k^{th}$ projection of the initial point $x_0$. The projection continues until convergence or when the terminating condition is satisfied. The main advantages of the parallel mode of the POCS method include the computational efficiency and the improved execution time.

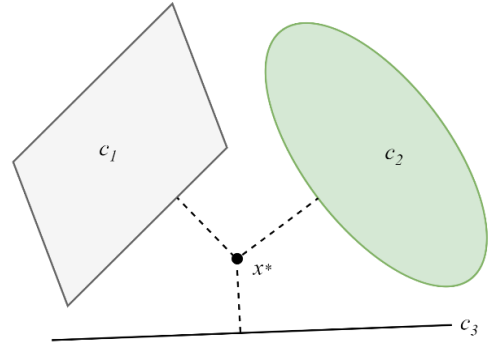

Fig. 2. Graphical interpretation of the parallel POCS for disjoint convex sets.

When the sets are disjoint, the parallel form of the POCS method converges to a point that minimizes the weighted sum of the squares of distances to the sets, which can be expressed as:

$$d = \sum_{i=1}^{n} w_i \|x^* - P_{c_i}(y^*)\|^2 \qquad (5)$$

where $x^*$ is the convergence point such that the distance $d$ defined by (5) is minimized. A graphical illustration of the convergence of the parallel POCS method is presented in Fig. 2.

### 2.4 The POCS-based Clustering Algorithm

For disjoint convex sets, the parallel mode of the POCS method converges to a point that minimizes the weighted sum of the squared distances. This property has been applied to clustering problems and the POCS-based clustering algorithm has been proposed in our preliminary work [11]. The POCS-based clustering algorithm considers each data point as a convex set and all data points in the cluster as disjoint convex sets. The objective function of the algorithm is defined as:

$$J = argmin \sum_{j}^{C} \sum_{i=1}^{n} w_i \|x_j - P_{c_i}(x_j)\|^2 \qquad (6)$$

$$w_i = \frac{\|x_j - d_i\|}{\sum_{p=1}^{n}\|x_j - d_p\|} \qquad (7)$$

with a constraint:

$$\sum_{i=1}^{n} w_i = 1 \qquad (8)$$

where $C$, $n$ represents the number of clusters and the number of data points in a cluster, respectively, while $P_{c_i}(x_j)$ is the projection of the cluster prototype $x_j$ onto the member point $d_i$ and $w_i$ denotes the weight of importance of the projection.

At first, the POCS-based clustering algorithm initializes $K$ cluster prototypes by adopting the prototype initialization method of the K-Means++ algorithm [12], then based on the Euclidean distance to the prototypes, each data point is assigned to one of the clusters which has the minimum distance from the data point. Until convergence, the algorithm computes new cluster prototypes using the equation:

$$x_{k+1} = x_k + \sum_{i=1}^{n} w_i(P_{c_i} - x_k), \qquad k = 0,1,2,\dots \quad (9)$$

with a constraint:

$$\sum_{i=1}^{n} w_i = 1 \qquad (10)$$

where $k$ is the iteration index. Starting from an initial point $x_0$, the projections converge to a point, $x_c$, that can minimize the objective function.

## 3. DATA PREPARATION

This section presents the preparation procedure of feature embedding data. In order to evaluate the robustness and applicability of the POCS-based clustering algorithm to complex clustering tasks including feature embedding clustering, various experiments on a variety of synthetic feature embedding datasets have been conducted. FaceNet [13] and autoencoder models [14] are applied to extract feature embeddings from the 5 Celebrity Faces [15] and MNIST [16] datasets for the clustering experiments.

### 3.1 FaceNet

FaceNet is a face recognition model proposed by researchers in Google in 2015 that has achieved state-of-the-art results on various face recognition benchmark datasets [13]. FaceNet can be utilized to extract high-quality face embedding features from face images, those face embeddings afterward can be used to train and develop a face recognition system.

The first dataset in this study is the 5 Celebrity Faces dataset [15]. This is a small dataset containing the photos of 5 celebrities: Ben Afflek, Elton John, Jerry Seinfeld, Madonna, and Mindy Kaling. The dataset is divided into training set and validation set. However, as this is a small dataset, we merge the two image sets to obtain a single image set of 118 images for clustering task. FaceNet afterward is adopted to extract face feature embeddings from all 118 images. The input and output shapes of FaceNet are 160x160x3 and 128x1, respectively. As the result, 118 face embeddings with the size of 128x1 are obtained and utilized as the input data for the clustering experiments. Several face image samples of Ben Affleck from the 5 Celebrity Faces dataset are shown in Fig. 3.

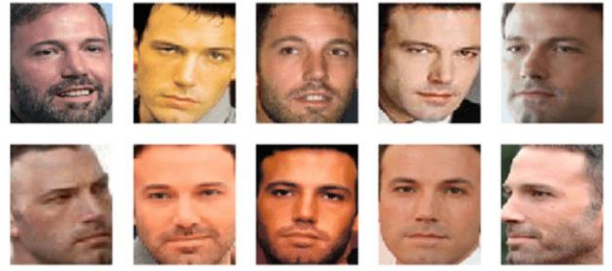

Fig. 3. Face image samples of Ben Affleck from the 5 Celebrity Faces dataset.

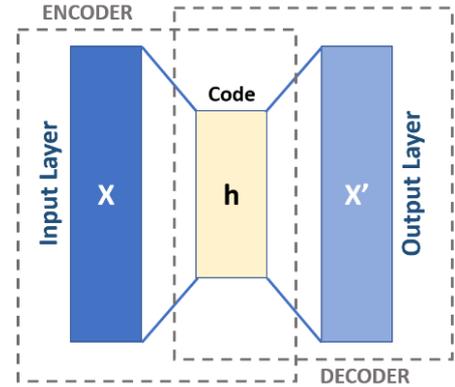

Fig. 4. Typical diagram of an autoencoder network.

### 3.2 Autoencoder (AE)

Autoencoder (AE) [14] is a type of artificial neural networks which is used to learn efficient embeddings of unlabeled data. A general AE model is comprised of 3 main parts: encoder, code, and decoder, as shown in Fig. 4. The input and output of an AE generally have the same shape. The encoder compresses the input data into a low-dimensional code (or representation) and the decoder reconstructs the code to produce the output data which is a copied version of the input. By training the network to perform a copying task, the codes or embeddings are optimized to learn the useful properties of the input data. Those embeddings then can be used for downstream tasks such as clustering and classification. The AE model has been widely applied to dimensionality reduction, feature extraction, image denoising, image compression, image search, anomaly detection, and single image haze removal [17][18][19]. Fig. 4. shows a typical diagram of an AE model.

For the sake of simplicity, in this paper, we adopt a simple AE model with a 4-layer encoder and a 4-layer decoder. We train the AE model on the MNIST dataset

```
Layer (type)                 Output Shape              Param #
=================================================================
input_4 (InputLayer)         [(None, 784)]             0
_________________________________________________________________
dense_6 (Dense)              (None, 128)               100480
_________________________________________________________________
dense_7 (Dense)              (None, 64)                8256
_________________________________________________________________
dense_8 (Dense)              (None, 32)                2080
=================================================================
Total params: 110,816
Trainable params: 110,816
Non-trainable params: 0
```

(a)

```
Layer (type)                 Output Shape              Param #
=================================================================
input_5 (InputLayer)         [(None, 32)]              0
_________________________________________________________________
dense_9 (Dense)              (None, 64)                2112
_________________________________________________________________
dense_10 (Dense)             (None, 128)               8320
_________________________________________________________________
dense_11 (Dense)             (None, 784)               101136
=================================================================
Total params: 111,568
Trainable params: 111,568
Non-trainable params: 0
```

(b)

Fig. 5. Description of the AE model used in this study: (a) encoder, and (b) decoder.

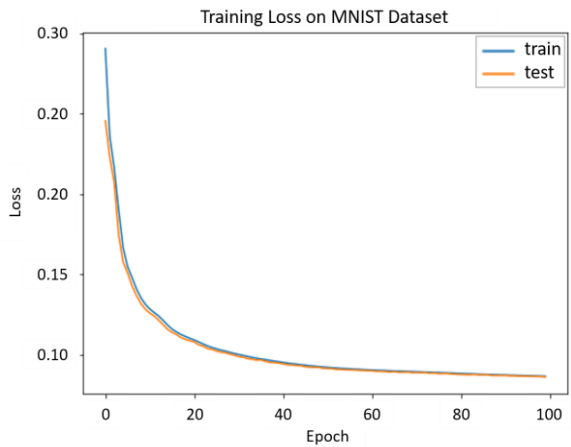

Fig. 6. Training curves of the used AE model on the MNIST dataset.

[16]. MNIST dataset contains 60,000 and 10,000 gray images for training and validation, respectively, with the image resolution of 28x28 pixels. We flatten all the images in the dataset to generate 784-d vectors which are utilized as input of the network. The embedding size is 32x1. The detailed description of the used AE model is shown in Fig. 5.

The processor used in the experiments was Intel(R) Core(TM) i7-4790K CPU @ 4.00GHz on a 64-bit operating system. We train the AE model in 100 epochs for each dataset. We apply the Adam optimizer [20] with a learning rate of 0.001 to train the network. The training curve (loss curve) is shown in Fig. 6 while several image reconstruction results are shown in Fig. 7. After the model converges, we extract the embeddings from the validation set for the clustering task. As the result, we obtain a set of 10,000 feature embeddings with the size of 32x1.

Table 1. Comparison between the POCS-based clustering and the K-Means++ algorithms in terms of clustering error on different feature embedding sets.

|       | K-Means++      | POCS-based     |
|-------|----------------|----------------|
| Face  | 111.2±0.3      | 111.3±0.7      |
| MNIST | 6,804.6±12.8   | 6,836.9±27.6   |

Table 2. Comparison between the POCS-based clustering and the K-Means++ algorithms in terms of execution time (ms) on different feature embedding sets.

|       | K-Means++     | POCS-based    |
|-------|---------------|---------------|
| Face  | 4.3±0.6       | 4.1±0.9       |
| MNIST | 916.9±301.8   | 771.2±269.6   |

## 4. EXPERIMENTS AND RESULTS

In order to evaluate the robustness and applicability of the POCS-based clustering algorithm to complex clustering tasks such as feature embedding clustering problems, various experiments and analyses are conducted. These experiments aim to thoroughly explain the convergence property of the POCS-based clustering algorithm on complex problems and compare its performances in terms of clustering error and execution speed with those of other classical clustering schemes including the K-Means and FCM algorithms. In all experiments, each algorithm is executed 20 times, then the mean and standard deviation values of clustering error and execution speed are measured and presented.

We conduct the comparison experiments under two conditions: same initial prototypes and different initial prototypes. Specifically, in the first condition, we compare the convergence performance of the POCS-based clustering algorithm with that of the K-Means algorithm when using the same initialized prototypes as these two algorithms have a similar working process. In this circumstance, the K-Means algorithm becomes the K-Means++ algorithm because we apply the prototype initialization procedure of the K-Means++ algorithm. On the other hand, in the second condition, we compare the performances of the POCS-based clustering, K-Means, and FCM algorithms, each algorithm is executed independently and the mean and standard deviation of clustering error and execution time are reported.

### 4.1 Condition 1: Same Initial Prototypes

Table 1 summarizes the performances of the POCS-based clustering and the K-Means++ algorithms in terms of clustering error on Face and MNIST embedding sets. As summarized in Table 1, the POCS-based clustering algorithm has similar performance compared with that of the K-Means++ algorithm in terms of clustering error. Even though the K-Means++ algorithm still shows a better result on MNIST embedding set by a minor gap.

The execution speeds of the two algorithms are also examined and reported in Table 2. As can be seen from Table 2, on the Face embedding set, the POCS-based clustering algorithm outperforms the K-Means++ algorithm in terms of the average convergence speed with a minor gap on the Face embedding set (4.1 ms compared with 4.3 ms, respectively) and a significant gap on MNIST feature embedding set (771.2 ms compared with 916.9 ms, respectively).

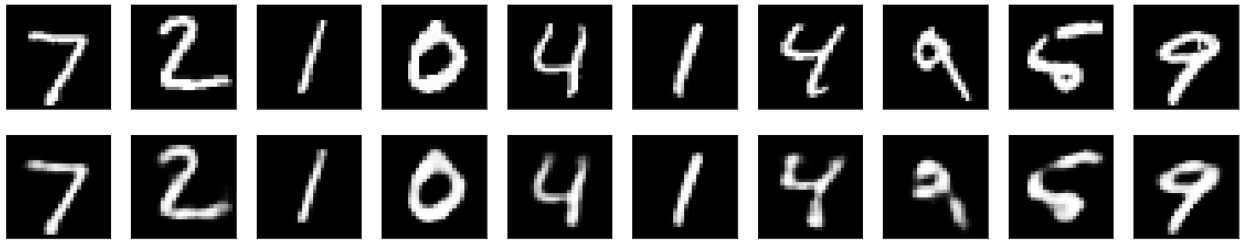

Fig. 7. Reconstructed images on MNIST dataset using the autoencoder model described in the paper (top: input image, bottom: reconstructed image).

Table 3. Comparison between the POCS-based clustering and the K-Means++ algorithms in terms of classification accuracy on different feature embedding sets.

|  | K-Means++ | POCS-based |
|---|---|---|
| **Face** | 99.36±1.7 | 99.53±1.5 |
| **MNIST** | 64.6±2.2 | 63.9±1.8 |

Table 4. Comparison of the K-Means, FCM, and POCS-based clustering algorithms in terms of clustering error on different feature embedding sets.

|  | K-Means | FCM | POCS-based |
|---|---|---|---|
| **Face** | 110.5±0.0 | 111.1±0.0 | 111.2±0.2 |
| **MNIST** | 6,692.4±0.1 | 8,323.0±0.0 | 6,721.5±15.3 |

Additionally, the classification accuracy of these two clustering algorithms is then investigated and shown in Table 3. As reported in Table 3, these two clustering algorithms share similar performances in terms of classification accuracy. The POCS-based clustering algorithm insignificantly outperforms the K-Means++ algorithm on the Face embedding set while the K-Means++ algorithm can marginally surpass the POCS-based clustering algorithm on the MNIST feature embedding set.

Based on the preceding results, we can empirically conclude that the POCS-based clustering algorithm can be considered a competitive method compared with the K-Means++ algorithm in feature embedding clustering tasks.

**4.2 Condition 2: Different Initial Prototypes**

Table 4 summarizes the performances of the K-Means, FCM, and POCS-based clustering algorithms in terms of clustering error. As can be seen from Table 4, the POCS-based clustering algorithm has a similar performance in terms of clustering error to that of the K-Means algorithm. While the FCM algorithm shows significantly higher clustering error.

Table 5. Comparison of the K-Means, FCM, and POCS-based clustering algorithms in terms of execution time (ms) on different feature embedding sets.

|  | K-Means | FCM | POCS-based |
|---|---|---|---|
| **Face** | 11.8±4.1 | 45.3±6.2 | 4.5±2.3 |
| **MNIST** | 282.1±30.8 | 1,484.9±40.2 | 615.9±243.7 |

Table 6. Comparison of the K-Means, FCM, and POCS-based clustering algorithms in terms of classification accuracy on different feature embedding sets.

|  | K-Means | FCM | POCS-based |
|---|---|---|---|
| **Face** | 100.0±0.0 | 93.2±0.0 | 99.2±2.2 |
| **MNIST** | 69.8±0.5 | 60.7±5.2 | 64.5±4.2 |

Table 5 summarizes the performances in terms of the execution time of the algorithms of interest. On the Face embedding set, the POCS-based clustering algorithm convincingly outperforms the other clustering schemes with 4.5±2.3 ms, which is 2.6 and 10 times faster than the K-Means and FCM algorithms. However, on the embedding set extracted from the MNIST dataset, the POCS-based clustering algorithm performs with the second-best results, yet still executes much faster than the FCM algorithm. The disadvantage of the POCS-based clustering algorithm here is the instability of the execution time. That is, depending on the initial prototypes, the algorithm may converge extremely fast or slow. Compared with the K-Means++ algorithm, the K-Means algorithm can converge much faster on the MNIST feature embedding set due to the difference in prototype initialization procedures. That is, the K-Means algorithm randomly picks the initial prototypes while the K-Means++ algorithm applies a careful seeding method for prototype initialization.

The performances of the examined algorithms in terms of classification accuracy are then investigated and shown in Table 6. As can be seen from Table 6, the POCS-based clustering algorithm and the K-Means algorithm have similar results on the Face embedding set, while the K-Means algorithm can produce a more favorable accuracy on the MNIST embedding set. On the other hand, the performance of the FCM algorithm can not surpass that of the other algorithms in our experiment.

As the result, the POCS-based clustering algorithm has a competitive performance when compared with that of the K-Means algorithm in feature embedding clustering problems. It implies that the POCS-based clustering algorithm has potential in a wide range of clustering tasks.

## 5. CONCLUSIONS

In this paper, the applicability of the POCS-based clustering algorithm, a novel clustering technique based on the projection onto convex set (POCS) method, to feature embedding clustering problems is examined. The POCS-based clustering algorithm applies the property of the POCS method to clustering problems and has been proved to be able to produce competitive performance compared to that of other classical clustering schemes in terms of clustering error and execution speed. In this study, the applicability of the POCS-based clustering algorithm to complex clustering tasks such as feature embedding clustering is further investigated. The experimental results on the embedding datasets generated from the 5 Celebrity Faces and MNIST datasets show that the POCS-based clustering algorithm can perform with favorable results and can be considered a promising tool for various data clustering tasks.